# Comparison between layer-to-layer network training and conventional network training using Deep Convolutional Neural Networks

Kiran Kumar Ashish Bhyravabhottla[1]   WonSook Lee[2]
[1,2]School of Electrical Engineering and Computer Science, Ottawa, Canada
[1,2]University of Ottawa

*Abstract*

*Convolutional neural networks have been widely deployed in almost all applications. It reached every boundary and scenario. Now, there has been significant development in neural architectures such as transfer learning, generative networks, diffusion models, and so forth. But each network's base is the convolutional neural architecture. In today's scenario, accuracy plays a crucial role. In general, accuracy mainly depends on the features. The features are extracted through the convolutional filters inside hidden layers. So, the layer in any architecture has a very vital role to play in the training process.*

*In this research, we propose a comparative analysis of layer-to-layer training and the conventional training of the network. In layer-to-layer training, the portion of the first layers is treated as a student network and the last layers are treated as a teacher network. During each step of training, the portions keep incrementing in the forward layers or student network and decrementing in the last layers or teacher network. This layer-to-layer comparison is tested on the VGG16, ResNext and DenseNet networks without using any pre-trained ImageNet weights and on a normal CNN model. The results are then compared with the conventional training method with VGG16 ResNext, DenseNet and the normal CNN model respectively.*

## 1. Introduction

Convolutional neural networks have gained momentum in image classification, object detection, and image segmentation applications. For certain real-world scenarios, traditional machine learning still has limitations despite its success and application in many practical applications. The problem, however, is that obtaining sufficient training data can be costly, time-consuming, or even impossible in many cases. This problem can be partially addressed by semi-supervised learning, which does not require mass-labeled data. For improved learning accuracy, semi-supervised approaches utilize a large amount of unlabeled data instead of a limited amount of labeled data. The resultant traditional models are usually unsatisfactory because unlabeled instances are also challenging to collect. Hence, transfer learning came into existence intending to transfer knowledge across domains with limited labeled data. In simple words, it is learning to transfer the generalization of experience. This creates a scenario of the ability to realize

---

https://github.com/ashish-AIML/LIII_lab

the situations through experiences. The commonly used transfer learning methodology is the ImageNet weights [2]. The idea of implementing ImageNet as pre-trained model weights is inspired by human beings' ability to transfer knowledge across domains. It leverages the knowledge from the source, i.e., ImageNet [2] data to improve the performance of the model and to minimize the number of labeled data required in the target domain.

Now, the main research focus is on improving accuracy. Critical applications required a very high amount of accuracy equaling nearly 99%. Hence, the growing intolerance of having accuracies less than even 97% is gaining momentum. This research is about exploring layer-to-layer training with a simple convolutional network as a teacher-student mechanism and analyzing its memory consumption, training speed, and performance with the normal conventional training methods.

## 2. Background and Motivation

Modern neural networks are composed of dozens or hundreds of layers that perform mathematical operations. These layers take a feature tensor as input and output activations corresponding to those features. The training algorithm iterates over a large dataset many times and minimizes the loss function. The full dataset is partitioned into mini-batches and iterated through the full dataset. This process is called an epoch. The training of a neural network consists of (a) forward pass, (b) backward pass, and (c) parameter synchronization. The forward pass (FP) analyses the model layer-by-layer in each iteration to determine the loss to the target labels and the loss function. GPU computing is needed for the forward and return passes. We determine the parameter gradients from the last layer to the first layer in the backward pass (BP) using the chain rule of derivatives for the loss [5]. We update the model parameters utilizing an optimization procedure, such as stochastic gradient descent, after each iteration (SGD) [4].

Since the datasets are complicated in today's scenario, several intense layer-based algorithms have been proposed to acquire higher accuracies. Many techniques such as optimizing parameters for existing algorithms to achieve better accuracies have been executed. In this research, we explore a different approach to training the neural network. Recent efforts have shown that front layers extract general features and deeper layers are more task-specific feature extractors [5]. Our research aims at exploring layer-wise training within a network without using any pre-trained residual network's weights [1] or any sort of pre-trained weights rather than training from the scratch.

## 3. Technical Approach

**Architecture:** Our architecture is a simple and normal convolutional neural network. It's a sequential training network. The base network is a *12-layered network*. The first layer is a 2D convolutional layer with *32* filters, each with a kernel size of *3x3*, *'same'* padding, and *ReLU* activation [7]. The input shape is the shape of a single image in the training data. The second layer is another 2D convolutional layer with *32* filters, also with a kernel size of *3x3*, *'same'* padding, and *ReLU* activation. The third layer is a 2D max pooling layer with a pool size of *2x2*. The fourth layer is a dropout layer with a rate of *0.25*, which randomly drops 25% of the inputs during training to prevent overfitting. The fifth layer is a 2D convolutional layer with *64* filters, each with a kernel size of *3x3*, *'same'* padding, and *ReLU* activation. The sixth layer is another 2D convolutional layer with *64* filters, also with a kernel size of *3x3*, *'same'* padding, and *ReLU* activation. The seventh layer is another 2D max pooling layer with a pool size of *2x2*. The eighth layer is another dropout layer with a rate of *0.25*. The ninth layer is a flattened layer that flattens the output of the previous layer into a 1D array. The tenth layer is a fully connected layer with *512* units and *ReLU* activation. The eleventh layer is another dropout layer with a rate of *0.5*. The final layer is another fully connected layer with *num_classes* units and a *softmax* activation function.

**VGG16 Architecture:** The first convolutional layer has *32* filters, each with a kernel size of *3x3*. So, the number of parameters in this layer is *(3 * 3 * input_channels + 1) * 32*, where *input_channels* is the number of channels in the input image (usually 3 for RGB images). In this case, the input shape is *(32, 32, 3)*, so the number of parameters in this layer is *(3 * 3 * 3 + 1) * 32 = 896*. The second convolutional layer has the same parameters as the first, so it also has *896* parameters. The max pooling layers and dropout layers do not have any parameters. The third convolutional layer has *64* filters, so the number of parameters in this layer is *(3 * 3 * 32 + 1) * 64 = 18496*. The fourth convolutional layer has the same parameters as the third, so it also has *18496* parameters. The first fully connected layer has *512* units, so the number of parameters in this layer is *(previous_layer_size + 1) * 512*, where *previous_layer_size* is the flattened size of the previous layer. In this case, the previous layer has a flattened size of *4096 (64 * 8 * 8)*, so the number of parameters in this layer is *(4096 + 1) * 512 = 2097664*. The second fully connected layer has *num_classes* units, so the number of parameters in this layer is *(previous_layer_size + 1) * num_classes*. In this case, the previous layer has a size of *512*, so the number of parameters in this layer is *(512 + 1) * num_classes*.

**ResNext Architecture:** The ResNext architecture takes an input tensor of shape specified by input_shape and produces an output tensor of shape which is number of classes, i.e., 100. The architecture consists of four groups of convolutional layers, each group containing two convolutional layers with the same number of filters. The number of filters is doubled in each group, starting from 64 in the first group. After each convolutional layer, batch normalization is performed, followed by the ReLU activation function. Max pooling is applied after each group to reduce the spatial size of the feature maps. The final layers of the network consist of a global average pooling layer followed by a fully connected layer with num_classes neurons and a softmax activation function to produce a probability distribution over the classes. This architecture is based on the ResNet architecture, which introduces residual connections to address the vanishing gradient problem in deep neural networks. However, the ResNext architecture extends ResNet by introducing a split-transform-merge strategy for the residual connections, which allows for more diverse representations to be learned by the network.

**DenseNet Architecture:** The architecture consists of a convolutional layer followed by batch normalization and ReLU activation, a series of dense blocks, and a global average pooling layer and a fully connected softmax output layer. Each dense block consists of a series of bottleneck layers and convolutional layers with concatenation of feature maps. The architecture ends with a global average pooling layer and a fully connected softmax output layer.

## 4. Experiments

We evaluate our model with the standard CIFAR100 [3] dataset with a normal CNN network and on dense networks such as VGG16 [6], ResNext [1], DenseNet [6] networks. For each training, the number of epochs is set to **300.** All the experiments are implemented on the Google Colab GPU notebooks. The performance metrics to evaluate the model are:

       (i)      total training time
       (ii)     accuracy
       (iii)    total memory consumption

### 4.1 Benchmark Datasets

**CIFAR100:** CIFAR-100 is a popular image classification dataset that contains 60,000 32x32 color images in 100 classes, with 600 images per class. The dataset is split into 50,000 training images and 10,000 testing images. The 100 classes in CIFAR-100 are grouped into 20 super classes, each containing five fine-grained classes.

### 4.2 Layer-to-Layer Training

The model is trained with layer-to-layer training. In this mechanism, we declare a single network with *'n'* layers. In the first step, the $1^{st}$ layer and the $(n-1)^{th}$ layer is trained, freezing the rest of the layers. Here, the 1st layer acts as a student network and the (n-1)th layer acts as a teacher network. In the second step, $2^{nd}$ layer is a student network and $(n-2)^{nd}$ layer is a teacher network that is trained and the rest layers are frozen. In the third step, $3^{rd}$ layer as a student network and $(n-3)^{rd}$ layer as a teacher network is trained, freezing the rest of the layers. This process is continued till *(n-i)* layers as *(i+1, (n-(i+1)), (i+2, (n-(i+2)), (i+3, (n-(i+3)),……… ((i+n/2), (n-(i+n/2)),* where 'n' is the number of layers, and 'i' is the student layers and value of *i=0*. After training with all the layer pairs, we perform an ensemble of all the layer pairs to print the final accuracy.

### 4.3 Standard Training

The performance is compared with the standard training of the networks. Normal training is a standard sequential training of all the layers at once.

### 4.4 Results and Discussion

In this section, we discuss the performance of layer-to-layer training compared with the standard training of both architectures respectively. The tabular comparison is shown in table 1.

**Standard CNN:** Since the accuracy factor plays an important role in critical applications, layer-to-layer training outperforms the standard training methods. As shown in Table 1, the accuracy of layer-to-layer training is **80%,** and that of standard layered training is **78%**. Crucial applications can utilize the layer-to-layer training method which is critical for accuracy. The other two performance metrics performed better than the layer-to-layer training. The total training time for standard training is **72.7 seconds** compared to **309.19 seconds** of layer-to-layer training. The total memory consumption of standard training is **5.27 GB** compared to **8.86 GB** of layer-to-layer training. Whenever the systems are RAM critical, standard layered training can be used, but with better RAM systems, layer-to-layered training is better since higher accuracies can be achieved.

**VGG16:** VGG16 [6] architecture is trained without pre-trained ImageNet weights but rather trained from scratch. The scenario of VGG16 is similar to that of the standard CNN. Accuracy is greater in layer-to-layer training than the standard training. But when compared with the standard CNN architecture, the accuracy of VGG16 in both training methods are almost negligible. Even after performing the ensemble method in the layer-to-layer training, the accuracy did not increase. The accuracy of layer-to-layer training is at **10%** and that of standard training is at **9.62%**. The total memory consumption of standard VGG16 is **5.4 GB** and that of layer-to-layer training is **6 GB**. The total training time of standard layers training is 265.5 seconds compared to that of 2311.3 seconds of layer-to-layer training.

**DenseNet:** Since the accuracy factor plays an important role in critical applications, layer-to-layer training outperforms the standard training methods. As shown in table 1, the accuracy of layer-to-layer training is **63.98%,** and that of standard layered training is **60.25%**. Crucial applications can utilize the layer-to-layer training method which is critical for accuracy. The other two performance metrics performed better than the layer-to-layer training. The total training time for standard training is **36045.163185596466 seconds** compared to **68081.966414779316 seconds** of layer-to-layer training. The total memory consumption of standard training is **5.26 GB** compared to **7.77 GB** of layer-to-layer training. Whenever the systems are RAM critical, standard layered training can be used, but with better RAM systems, layer-to-layered training is better since higher accuracies can be achieved.

**ResNext:** ResNext architecture is trained without pre-trained ImageNet weights but rather trained from scratch. Accuracy is greater in layer-to-layer training than the standard training Even after performing the ensemble method in the layer-to-layer training, the accuracy did not increase. The accuracy of layer-to-layer training is at **56.85%** and that of standard training is at **55.28%**. The total memory consumption of standard training of ResNext is **4.9 GB** and that of layer-to-layer training is **6.9 GB**. The total training time of standard layers training is **3930.0629668235779 seconds** compared to that of **8790.185438156128 seconds** of layer-to-layer training.

The second half of the network is chosen as a teacher model since it is processed with intense filter sizes hence more knowledge can be extracted from these layers. It is generally believed that the last layers typically involve global pooling operations and fully connected layers act as classifiers. These layers are responsible for extracting a high-level feature from the input images, which can be used to classify the image into different classes. Therefore, in this sense, the last layers of a CNN can be considered global feature extractors, as they take into account the entire image and produce a summary of its features that can be used for classification. Hence, they are more knowledgeable than the initial layers. On the other hand, the earlier layers of a CNN typically perform local feature extraction by detecting low-level visual features such as edges, corners, and textures in different regions of the input image. These features are gradually combined and transformed by subsequent layers to form higher-level features that are increasingly global. Hence, accuracy is greater in layer-to-layer training than the standard layer training.

| Network Architecture | Training Method | Total Training Time | Accuracy | Total Memory Consumption |
|---|---|---|---|---|
| Standard CNN | Layer-to-Layer training | 309.1986117362976 seconds | **80%** | 8862.59765625 MB |
| | Standard training | **72.70218634605408 seconds** | 78% | **5277.4296875 MB** |
| VGG16 | Layer-to-Layer training | 2311.3207755088806 seconds | **10%** | 6035.421875 MB |
| | Standard training | **265.588809967041 seconds** | 9.62% | **5436.65234375 MB** |
| DenseNet | Layer-to-Layer training | 68081.966414779316 seconds | **63.98%** | **7.77 GB** |
| | Standard training | **36045.163185596466 seconds** | 60.25% | 5.2 GB |
| ResNext | Layer-to-Layer training | 8790.185438156128 seconds | **56.85%** | 6.9 GB |
| | Standard training | **3930.0629668235779 seconds** | 55.28% | **5.5 GB** |

Table 1. Performance metrics of training methods on the CIFAR100 dataset

## 5. Conclusion and Future Work

We have researched layer-to-layer training within the same network and shown its advantage by comparing it with normal standard training. The second section of the architecture is treated as teacher network because of its ability to extract dense features. Layer-to-layer training resulted on greater accuracies for both normal CNN and transfer learning architectures. As the epochs increased, the accuracy increases in layer-to-layer training. Although, the layer-to-layer training resulted in higher accuracies, its training speed and memory consumption is higher compared to the normal conventional training. But as the dataset quantity increases, the RAM consumption for layer-to-layer training increases drastically and requires more than 15 GB of RAM for denser architectures such as ResNext V1 and ResNext V2. Therefore, more RAM requirement is recommended for layer-to-layer training. We can conclude that as the dataset classes increases and number of layers increases, more hyper-tuning is required, and pre-trained weights are required to improve the accuracy.

In the future, the dense architectures can be trained with higher RAM such as multi-GPUs and train with much higher datasets to see the variation of accuracy between small datasets and very large datasets. Further, instead of computing multi-training methods and increasing the number of epochs consuming memory, we can deploy a layer-to-layer method to improve the performance of the network resulting in higher accuracy. In future, the experiments can be based on increasing the layers, epochs and training with high quality and quantity datasets to evaluate the performance of layer-to-layer training. Further, the experiments can be conducted with multi-GPUs and multi-threading to check the training speed of both normal training and layer-to-layer training. Since, the layer-to-layer training resulted in better accuracy than the normal conventional training, the future work can even focus on extending this methodology to object detection and image segmentation models to evaluate the performance.